\pdfoutput=1
\documentclass[11pt]{article}
\usepackage[]{emnlp2021}
\usepackage{times}
\usepackage{latexsym}

\usepackage{microtype}

\usepackage{array}
\usepackage{multirow}
\usepackage{graphicx}
\usepackage{subfigure}
\usepackage{array, diagbox}

\newcommand{\revised}[1]{\textcolor{black}{#1}}

\title{CR-COPEC: Causal Rationale of Corporate Performance Changes \\to Learn from Financial Reports}

\author{\begin{tabular}[c]{@{}c@{}}Ye Eun Chun$^{1}$\thanks{~~Both authors contributed equally to this work.}, Sunjae Kwon$^{2*}$, Kyunghwan Sohn$^{3}$, Nakwon Sung$^{4}$, \\ Junyoup Lee$^{5}$, Byungki Seo$^{1}$, Kevin Compher$^{6}$, Seung-won Hwang$^{4}$, Jaesik Choi$^{3,7}$\thanks{~~Corresponding author: jaesik.choi@kaist.ac.kr}\end{tabular}\\
\begin{tabular}[c]{@{}c@{}} 
  $^{1}$UNIST, $^{2}$UMass Amherst, $^{3}$KAIST, $^{4}$SNU\\$^{5}$ Ajou University, $^{6}$ Foresight Institute, $^{7}$ INEEJI
\end{tabular}\\
  }
    
\date{}

\begin{document}
\maketitle
\begin{abstract}
In this paper, we introduce CR-COPEC called \textit{\textbf{C}ausal \textbf{R}ationale of \textbf{Co}rporate \textbf{Pe}rformance \textbf{C}hanges} from financial reports. This is a comprehensive large-scale domain-adaptation causal sentence dataset to detect financial performance changes of corporate. CR-COPEC contributes to two major achievements. First, it detects causal rationale from 10-K annual reports of the U.S. companies, which contain experts’ causal analysis following accounting standards in a formal manner. This dataset can be widely used by both individual investors and analysts as material information resources for investing and decision-making without tremendous effort to read through all the documents. Second, it carefully considers different characteristics which affect the financial performance of companies in twelve industries. As a result, CR-COPEC can distinguish causal sentences in various industries by taking unique narratives in each industry into consideration. We also provide an extensive analysis of how well CR-COPEC dataset is constructed and suited for classifying target sentences as causal ones with respect to industry characteristics. Our dataset and experimental codes are publicly available\footnote{\url{https://github.com/CR-COPEC/CR-COPEC}}.
\end{abstract}

\section{Introduction}
Many critical decisions on events may require appropriate explanations of decisions based on accurate causal rationale. Justifying the root statements is directly related to identifying the causes of the events. When one could observe an event that is an apparent cause of a desired outcome, one can make a proper decision with confidence on events.

There has been extensive research in extracting causes of events in numerical data. As an example, Granger causality finds linear temporal dependence between two (or more) temporal sequences \cite{Granger}. Shapley values derive the contributions of individual input attributes when a decision is made by a complex function \cite{ShapleyLloyd1971Cocg}. These methods can explain numerical causes of decisions made by automated systems, e.g. Robo-advisers in financial services \cite{Pmlr-v48-hwangb16, Forbes, chhatwani2022does}. However, human uses various types of information including numerical and textual inputs when making an important decision. As an example, analysts write summarized reports by extracting related causal information from multiple textual sources in conference calls, annual reports, earning statements and markets reports. 

Our research goal is to extract causal rationales from financial reports. In general, when investing firms predict a certain financial performance, they provide analysts' reports to support their predictions. Thus, we want to generate appropriate explanations for performance changes of a corporate from official documents. Our algorithm classifies causal sentences from documents and provides binary classification results.

We consider fine-tuning of Pre-trained Neural Language Models (NLMs) for the causality modeling tasks. Pre-trained NLMs have been state-of-the-arts for many Natural Language Processing (NLP) tasks. For example, NLMs such as BERT \cite{DBLP:journals/corr/abs-1810-04805} and ALBERT \cite{lan2019albert} demonstrate outstanding performance in some tasks such as answering questions \citep{clark2020electra, suissa2023question} and computing conditional probabilities of masked words in a sentence \citep{kwon2022medjex}. \revised{Nonetheless, recent research indicates that the size of human-annotated data continues to be a significant factor influencing the performance of models \citep{gu2022ppt, mehrafarin2022importance}.} 

There are previous works that introduce dataset for causality detection in the financial domain \citep{el2016learning, mariko2020financial}. However, existing studies lacked consideration of industry-specific characteristics. Thus, it can be beneficial to consider it because the items in the financial statements that greatly affect financial performance are different for each company's primary business.

Our main contribution is to collect sufficient annotations to achieve reasonable causality detection performance with NLMs. We achieved this by collecting over 283K sentences from 1,584 10-K annual reports, that give a detailed summary of the financial status and business operations of each company, along with audited financial statements. Then, we manually label individual sentences whether the sentences explain the cause of certain financial performance changes. We name the 283K pairs of a sentence and a corresponding label as \emph{Causal Rationale of Corporate Performance Changes} (CR-COPEC). CR-COPEC is built on a large scale with guides of experts in the financial domain. Trained with our dataset, BERT can distinguish sentences containing main causes of its financial events, from annual reports written officially from most U.S. public companies. Thus, individual investors can save efforts to read a huge amount of reports by themselves. 

However, we find collecting the dataset does not solve the problem as itself. One challenge we observe in the process of annotation is \textbf{diverse causality} in industries. That is, we find diverse causalities over different sectors. Another challenge is \textbf{imbalanced training}, where the number of data for each sector varies. 
Thus, we need to build a model carefully, as applying a common model to all sectors does not work in our problem setting. We provide extensive analysis on our CR-COPEC dataset to overcome these issues.

\section{Related Work}

The causal rationale is ``the true sufficient rationales to fully predict and explain the outcome without supurious information.'' \citep{zhang2023towards} and extracting rationale is invaluable when a decision has to be made. Research on extracting rationale from text has been tried with various types of text documents \citep{blanco2008causal, ittoo2011extracting, lu2022rationale}. A model detecting and identifying rationale from chat messages was suggested in \cite{Alkadhi}. Bug reports from Chrome web browser were used as main sources to extract rationale \cite{RogersB2012Etfr}. Moreover, patent documents were utilized to discover design rationale \cite{LiangYan2012LtWD}. To extract causal textual structures, one may consider a rule-based system where specific words such as `due to', `owing to' and `affects' are listed to identify sentences including causal information for prediction \cite{girju2002text, CHANG2006662,Sakai}.

Diverse studies have explored causality extraction datasets \citep{xu2020review, ali2021causality, yang2022survey}. Some widely recognized datasets include SemEval-2007 task 4 \citep{girju2007semeval}, SemEval-2010 task 8 \citep{hendrickx2019semeval}, PDTB 2.0 \citep{prasad2008penn}, BioInfer \citep{pyysalo2007bioinfer}, and ADE \citep{gurulingappa2012development}. Among these, three datasets stem from general domain, while two originate from the biomedical field (BioInfer and ADE). The maximum number of causality sizes of these datasets is less than 10K (PDTB 2.0, 9,190 causal examples). In contrast, CR-COPEC contains a larger substantial data size of 105,861 sentences, including 11,132 sentences with causal rationales. In addition, CO-COPEC targets the financial domain especially focused on corporate performance changes from financial reports. These characteristics contribute to the uniqueness of CR-COPEC and potentially helpful for the decision of financial experts.

There are previous studies that introduce causal rationale corpus in the financial domain. To identify causal sentences from UK Preliminary Earning Announcements (PEAs), thirteen performance keywords including `sales', `revenue' and `turnover' are used and selected sentences are annotated by human \citep{el2016learning}. \citet{mariko2020financial}'s study is the most related our research. Herein, the authors built FinCausal corpus collected from financial news and websites which labeled with tags indicating the presence of causality and causal chunk as quantitative or non-quantitative. The corpus consists of two subtasks. The first subtask (FinCausal Task 1) is a binary classification task that targets to extract text having causality. The second subtask (FinCausal Task 2) is a relation extraction task identifying the substrings that indicate cause and effect. The main difference between CR-COPEC and FinCausal is that texts are written with a formal tone in 10-K reports since they should comply with regulatory rules. In addition, CR-COPEC solely concentrates on causal rationales of accounting items considering unique characteristics of various industries. Meanwhile, FinCausal dataset itself is written with a casual tone in general because it is collected from news or web contents. The detailed comparison of two datasets is described in Appendix~\ref{apx:dataset_comparison}





\section{Causal Rationale of Corporate Performance Changes Dataset}
\label{sec:corpora_construction_process}
\begin{figure}[]
\centering
\includegraphics[height=0.8\linewidth]{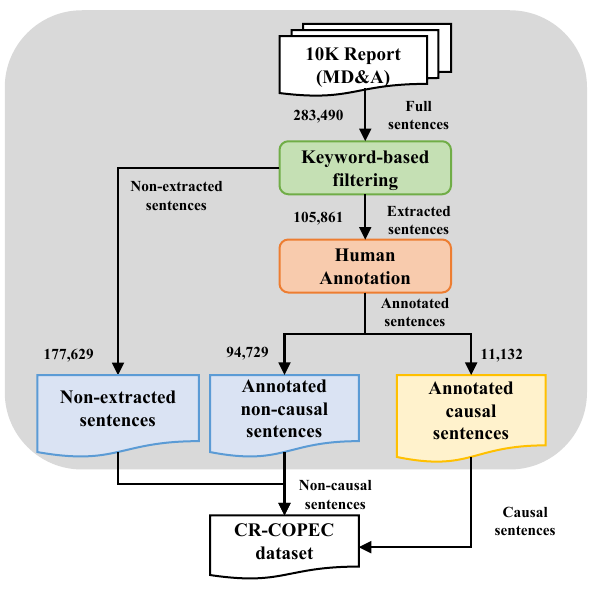}
\setlength{\belowcaptionskip}{-10pt}
\vspace*{-3mm}
\caption{Overview of CR-COPEC dataset construction process.}
\label{fig:CR-COPEC}
\end{figure}

\revised{Our goal is to collect causal rationale of sentences that  contain causal rationales in predicting changes in corporate performance based on key accounting items within the target industrial sector. For this purpose, we target Management’s Discussion and Analysis (MD\&A) section of 10-K reports since it provides the company's perspective on its operations and financial results of the prior fiscal year. We gather MD\&A reports filed in 1997 and 2017. MD\&A of 1997 is gathered from the existing MD\&A data repository \cite{Kogan:2009:PRF:1620754.1620794} and 2017's is downloaded directly from the Securities and Exchange Commission (SEC) system. Then, we construct CR-COPEC dataset through keyword-based filtering (Section~\ref{sec:keyword_based_filtering}) and human annotation process (Section~\ref{sec:human_annotation}) illustrated in Figure~\ref{fig:CR-COPEC}. The validity of the dataset is verified in Section~\ref{sec:corpus_validation}. In addition, we provide protocols of the dataset to train NLMs in Section~\ref{sec:Dataset to Validate Training Performance}.}

\begin{table*}[!ht]
    \centering
    \captionsetup{justification=centering}

    \resizebox{\textwidth}{!}{\begin{tabular}{c|c|c|c|c|c|c|c|c|c|c|c|c|c}
    \hline
    \textbf{} &  
    \multicolumn{1}{|c|}{\textbf{1}} & 
    \multicolumn{1}{|c|}{\textbf{2}} & 
    \multicolumn{1}{|c|}{\textbf{3}} & 
    \multicolumn{1}{|c|}{\textbf{4}} & 
    \multicolumn{1}{|c|}{\textbf{5}} & 
    \multicolumn{1}{|c|}{\textbf{6}} & 
    \multicolumn{1}{|c|}{\textbf{7}} & 
    \multicolumn{1}{|c|}{\textbf{8}} & 
    \multicolumn{1}{|c|}{\textbf{9}} & 
    \multicolumn{1}{|c|}{\textbf{10}} & 
    \multicolumn{1}{|c|}{\textbf{11}} & 
    \multicolumn{1}{|c|}{\textbf{12}} & 
    \multicolumn{1}{|c}{\textbf{Total}} \\ 
    \hline
    \hline
    Causes & \begin{tabular}[c]{@{}c@{}}1,145\\(5.70\%)\end{tabular}& \begin{tabular}[c]{@{}c@{}}346\\(4.55\%)\end{tabular}& 
    \begin{tabular}[c]{@{}c@{}}1,528\\(4.36\%)\end{tabular}& 
    \begin{tabular}[c]{@{}c@{}}497\\(3.20\%)\end{tabular}& 
    \begin{tabular}[c]{@{}c@{}}283\\(5.62\%)\end{tabular}& 
    \begin{tabular}[c]{@{}c@{}}2,505\\(4.19\%)\end{tabular}& \begin{tabular}[c]{@{}c@{}}721\\(3.24\%)\end{tabular}& \begin{tabular}[c]{@{}c@{}}490\\(2.46\%)\end{tabular}& 
    \begin{tabular}[c]{@{}c@{}}1,424\\(4.30\%)\end{tabular}& 
    \begin{tabular}[c]{@{}c@{}}563\\(2.80\%)\end{tabular}& 
    \begin{tabular}[c]{@{}c@{}}558\\(2.18\%)\end{tabular}& 
    \begin{tabular}[c]{@{}c@{}}1,072\\(5.51\%)\end{tabular}& \begin{tabular}[c]{@{}c@{}}11,132\\(3.93\%)\end{tabular}\\
    \hline
    Non-Causes     & 18,959 & 7,255 & 33,485 & 15,054 & 4,750 & 57,292 & 21,538 & 19,398 & 31,698 & 19,529 & 25,014 & 18,386 & 272,358 \\
    \hline
    Total              & 20,104 & 7,601 & 35,013 & 15,551 & 5,033 & 59,797 & 22,259 & 19,888 & 33,122 & 20,092 & 25,572 & 19,458 & 283,490\\
    \hline
    \textbf{\# of Doc} & \textbf{140} & \textbf{61} & \textbf{248} & \textbf{79} & \textbf{36} & \textbf{379} & \textbf{76} & \textbf{55} & \textbf{138} & \textbf{127} & \textbf{119} & \textbf{126} & \textbf{1,584} \\
    \hline
    \end{tabular}}
        \vspace*{-2mm}
        \caption{Dataset (CR-COPEC) Composition: number of sentences and documents in each sector \\(The ratio of causes to total sentences is in parenthesis).}
        \setlength{\belowcaptionskip}{-10pt}
        \vspace*{-3mm}
            \label{sectiontable}
\end{table*}

\begin{table*}[!ht]
    \footnotesize
    \centering
    \begin{tabular}{p{2\columnwidth}}
    \hline
    \textbf{Sector of Industry} - 
    {{Example}} {{[Document]}} \\
    \hline
    \hline
    \textbf{Consumer Durables} - 
    The sales increase for fiscal 1996 was principally due to improved sales of buses and ambulances. [Collins Industries, Inc., January, 1997]\\
    \hline
    \textbf{Manufacturing} -
    The increase in 1996 net sales was due primarily to increases in sales revenues recognized on the contracts to construct the first five Sealift ships, the Icebreaker and the forebodies for four double-hulled product tankers, which collectively accounted for 63\% of the Company's 1996 net sales revenue. [Avondale Industries, Inc., March, 1997]\\
    \textbf{Energy} - 
    Gas revenue increased \$32.9 million or 81\% because of a 39\% price increase combined with a 30\% increase in production. [Cross Timbers Oil Co., March, 1997]\\
    \hline
    \textbf{Chemicals} - 
    Loss of margin was principally due to sales price decreases and raw material price increases in the pyridine and related businesses, and higher manufacturing costs due to weather related problems in the first quarter 1994. [Cambrex Corp., March, 1997]\\
    \hline
    \textbf{Finance} - 
    Mortgage investment income decreased for 1995 as compared to 1994 primarily due to the assignment to HUD of the mortgage on El Lago Apartments in June 1995. [American Insured Mortgage Investors Series 85 L P, March, 1997]\\
    \hline
    \end{tabular}
        \vspace*{-2mm}
        \caption{Examples of causal rationale of sentences by each sector.}
        \setlength{\belowcaptionskip}{-13pt}
        \vspace*{-5mm}
            \label{explainableexample}
    
\end{table*}

\subsection{Keyword-based Filtering}
\label{sec:keyword_based_filtering}
Since sentences we aim for detecting are rare in a single document, it is costly expensive to manually annotate every collected sentence. To mitigate the cost issue, previous studies extract text containing keywords such as domain specific terminologies \citep{el2016learning, fonseca2023assessing} or causal phrases \citep{ Sakai, durlich2022cause}.
In our work, we utilize keyword-based filtering to extract candidate causal sentences.
In particular, we form a list of keywords\footnote{1) result from, 2) result of, 3) due to, 4) cause, 5) impact, 6) because of, 7) increase, 8) decrease, 9) decline, 10) negative, 11) contract and 12) significant} including causal trigger phrases of changes in financial performance. We filter out approximately 62.7\% \textit{non-extracted sentences} (177,629 out of 283,490) and simultaneously extract the remaining 37.3\% \textit{extracted sentences} (105,861 out of 283,490) from the total sentences of the MD\&A section. 
Section~\ref{sec:corpus_validation} discusses on the coverage of the keyword list.

\subsection{Human Annotation Process}
\label{sec:human_annotation}


\subsubsection{Data Annotation following SIC}
\label{sec:industrial_categories}
We describe the process of analyzing causal rationale of sentences by considering companies' Standard Industrial Classification (SIC) that classifies the U.S. companies according to their primary business.\footnote{\url{https://www.sec.gov/info/edgar/siccodes.htm}} For this, we divide 10-K reports into twelve industries which are predefined with respect to the SIC codes \citep{french}. These twelve categories include 1) consumer non-durables, 2) consumer durables, 3) manufacturing, 4) energy, 5) chemicals, 6) business equipment, 7) telephone, 8) utilities, 9) shops, 10) health, 11) finance and 12) others. Note that, throughout the paper, sectors are numbered in this order.


Since major factors that affect the financial performance of a company in each industry can be different, we build an annotation guideline for each sector by taking different accounting items among industries into consideration. Under the supervision of a finance faculty at the school business administration, we craft the annotation guideline by considering the main items of the balance sheets and income statements since those items are closely related to financial performance of companies. 
Then, we proceed annotation according to the annotation guideline. Each MD\&A document is randomly assigned to an annotator. The number of documents and sentences from each sector is reported in Table~\ref{sectiontable}. Examples of sentence containing main factors mostly observed in each sector are shown in Table~\ref{explainableexample}. Additional examples of each sector's main factors are demonstrated in Appendix~\ref{sec:main_factors} and causal/non-causal rationale of sentences can be found in Appendix~\ref{sec:appendix_additional_examples_of_causal_non_causal_sentences}.

\subsubsection{Annotator Sensitivity}
\label{sec:annotation_sensitivity}

We distinguish annotators into general and unskilled annotators based on the proficiency of the guideline.
We regard annotators who 1) 
participated in the development of the annotation guideline and 2) labeled more than 30K sentences as general annotators, otherwise unskilled annotators. 
The general annotators label each extracted sentence as non-causal or causal one. 100,046 of the sentences were tagged by two general annotators.

We regard labels annotated by general annotators as a standard and train a model with these labels alone. We call this model as an initial teacher model. Then, we apply this teacher model to documents that are labeled from unskilled annotators. If all labels in a single MD\&A document match with predictions of the teacher model, we add them to the previous training set. We train another teacher model with the new version of training set again and apply this model to the rest of other documents. We repeat this process until no matched document is found. As a result, we collect 1,584 10-K reports and 105,861 sentences containing 11,132 causal and 94,729 non-causal sentences.

Note that, all the sentences from the same document are assigned to the same annotator, and the annotators label the current sentence based on its previous context. Details on the annotation environment of annotators can be found in Appendix~\ref{sec:annotation_env}.

\begin{table}[]
\small
\centering
\begin{tabular}{l|cc|c}
\hline
             & Causal & Non-causal & Overall \\
\hline
\hline
Extracted     & 38     & 347        & 385     \\
Non-extracted & 6      & 677        & 683     \\
\hline
Overall      & 44     & 1,024      & 1,068  \\
\hline
\end{tabular}
\vspace*{-2mm}
\caption{The result of the coverage analysis of causal sentences extracted by the keywords.}
\setlength{\belowcaptionskip}{-15pt}
\vspace*{-5mm}
\label{tab:rule_coverage}
\end{table}
\subsection{The Validity of the Dataset}
\label{sec:corpus_validation}
In this section, we discuss the validity of our dataset construction process; 1) Keyword-based filtering and 2) Human Annotation.
\subsubsection{Verifying Keyword-based Filtering}
We randomly sample a small number of sentences from full sentences and classify causal sentences manually. Table~\ref{tab:rule_coverage} is the result of analyzing the coverage of keywords. From 1,068 sampled sentences, 385 sentences are extracted with the keyword list, and the rest of 683 sentences are filtered out during the keyword-based filtering process. Out of extracted sentences, 38 sentences are annotated as causal one. Meanwhile, six sentences are turned out to be causal rationale of sentences from non-extracted sentences. From these results, we can see that the keywords can be an effective indicator of causal rationale of sentences showing high recall performance of 86.4\% (38 out of 44). The ratio of causal rationale of sentences and the non-extracted sentences are 9.8\% (38 out of 385) and 0.9\% (6 out of 683) respectively.

\subsubsection{Verifying Human Annotation Process}
\label{sec:veryfying_annotation}
\begin{figure*}[!ht]
\centering
\begin{center}
\includegraphics[width=\linewidth]{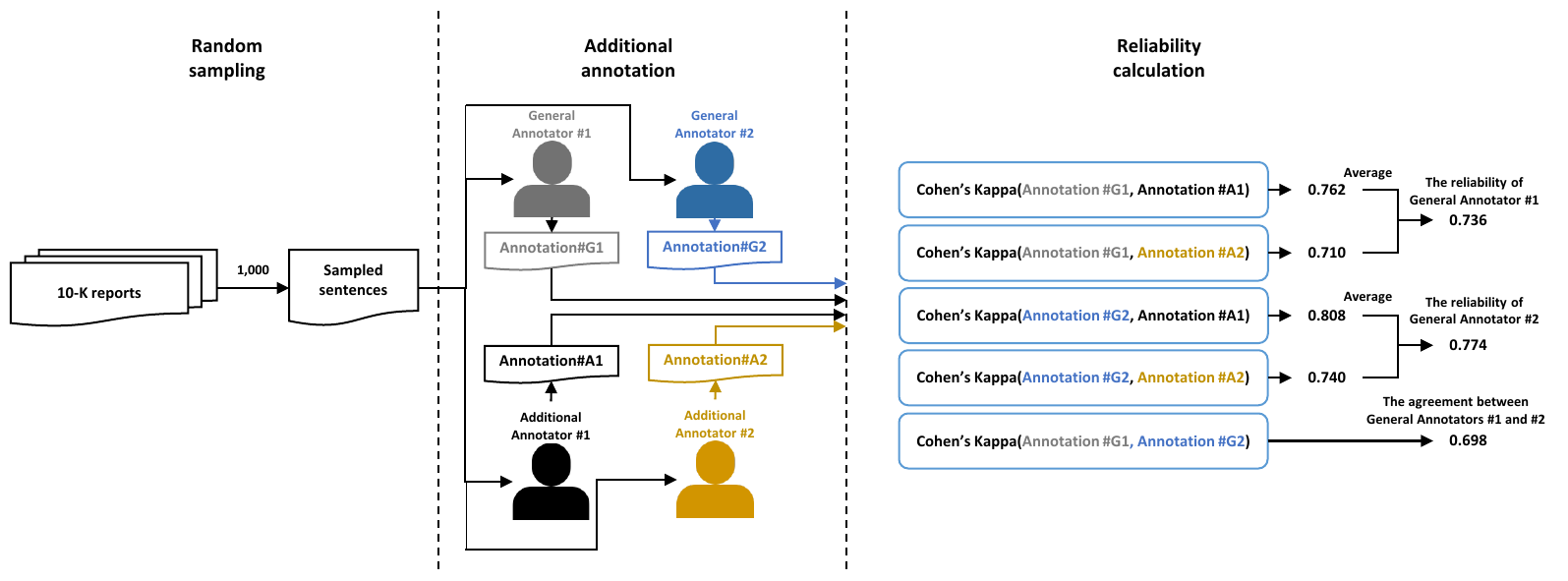}
\end{center}
\setlength{\abovecaptionskip}{-3pt}
\caption{Overview of the verifying the human annotations of general annotators.}
\vspace*{-3mm}
\label{fig:verify_human_annotation}
\end{figure*}

\begin{table*}[!ht]
    \small
    \centering
    \captionsetup{justification=centering}

    {\begin{tabular}{c|c|c|c|c|c|c|c|c|c|c|c|c}
    \hline
    \textbf{}  & 1 & 2 & 3 & 4 & 5 & 6 & 7 & 8 & 9 & 10 & 11 & 12 \\
    \hline
    \hline
    Train/Valid  &5.71	&4.58	&4.38	&3.18	&5.68	&4.18	&3.24	&2.45	&4.29	&2.80	&2.19 &5.53 \\
    \hline
    Test     &5.56	&4.37	&4.25	&3.34	&5.23	&4.24	&3.28	&2.58	&4.33	&2.85	&2.08	&5.38 \\
    \hline
    \end{tabular}}
        \caption{Total Dataset Causal Composition: the ratio of causal sentences in each sector (\%).}
        \setlength{\belowcaptionskip}{-12pt}
        \vspace*{-3mm}
            \label{causalratiotable}
\end{table*}

Human annotation task is sensitive to the reliability of the general annotators. Thus, to verify the reliability between annotators, we conduct the following steps. To be specific, this process includes three phases: 1) Random sampling, 2) Additional annotation and 3) Reliability calculation. Figure~\ref{fig:verify_human_annotation} is a conceptual diagram for the process. 

To begin with, we randomly sample 1,000 sentences from 10-K reports. Then, we ask for the general annotators and two additional annotators \footnote{Herein, additional annotator \#1 is working in the computational linguistic domain for more than 6 years and additional annotator \#2 is a financial expert who worked at Financial Supervisory Service for more than 8 years. Note that, both general annotators are $\kappa > 0.7$, when compared to additional annotator \#2, indicating that the annotations of the general annotators are reliable.} to annotate the sampled sentences. In this case, those additional annotators were trained on the annotation guideline and corpus labeling. As a result, we can get annotation \#G1, and annotation \#G2 from the general annotators and annotation \#A1, and annotation \#A2 from the additional annotators. 

Finally, we calculate the reliability of the sampled sentences of the general annotators \#1 and \#2. 
For this purpose, we use Cohen's Kappa \citep{cohen1960coefficient}, which evaluates the inter reliability of two annotators.
As a result, we observe that the agreements of the general annotator \#1 are 0.762, and 0.710 from the additional annotators \#1 and \#2, respectively, being able to be interpreted as a substantial agreement ($\kappa > 0.6$) \citep{viera2005understanding}. Furthermore, the agreements of the general annotator \#2 are 0.808, and 0.740 from the additional annotators \#1 and \#2, respectively. Finally, the agreement between general annotators is 0.698. 

\subsection{Dataset to Validate Training Performance}
\label{sec:Dataset to Validate Training Performance}

This section describes how we recompose CR-COPEC dataset as suitable to train NLM. In addition, we provide protocols for different version of datasets through Section~\ref{sec:Protocols for Total Dataset} and Section~\ref{sec:Protocols for Fraction of Dataset}.

\subsubsection{Protocols for Total Dataset}
\label{sec:Protocols for Total Dataset}

In CR-COPEC dataset, the ratio of causal to non-causal is imbalanced, the robustness of learning causality can be affected by the ratio of causal in each train/valid/test dataset. Thus, we carefully select train, valid and test dataset so that each sector composes similar causal ratio between different datasets. In this process, we iterate random selection until the ratio difference is less than 0.5\%. The proportion of train/valid/test follows a 81\%/9\%/10\% of total CR-COPEC dataset. The causal composition of train/valid/test dataset for each sector is reported in Table~\ref{causalratiotable}.

\subsubsection{Protocols for Fraction of Dataset}
\label{sec:Protocols for Fraction of Dataset}
In Section~\ref{sec:sectorspecific}, we compare cross-sector performances to figure out transferable information between sectors. As we observe in Table~\ref{sectiontable}, the size of each sector varies from 5K to 60K. Thus, in order to control the effect of size on performance, we randomly sample 3,500 sentences from each sector of train dataset. Since the smallest number of train datasets among all sectors is less than 4,000.

\section{Analysis on CR-COPEC Dataset}

\begin{table*}[!ht]
\scriptsize
\centering
\begin{tabular}{cc|cccccccccccc|c}
\hline
                                                                                                                   &    & \multicolumn{12}{c|}{Sector Test Dataset}                                                                                                                                                                                                                                                                                                                                                                                                        & \multirow{2}{*}{Avg.} \\ \cline{3-14}
                                                                                                                   &    & \multicolumn{1}{c|}{1}              & \multicolumn{1}{c|}{2}              & \multicolumn{1}{c|}{3}              & \multicolumn{1}{c|}{4}              & \multicolumn{1}{c|}{5}              & \multicolumn{1}{c|}{6}              & \multicolumn{1}{c|}{7}              & \multicolumn{1}{c|}{8}              & \multicolumn{1}{c|}{9}              & \multicolumn{1}{c|}{10}             & \multicolumn{1}{c|}{11}             & 12             &                       \\ \hline\hline
\multicolumn{1}{c|}{\multirow{12}{*}{\begin{tabular}[c]{@{}c@{}}Sector-only\\ Same-size \\ Model\end{tabular}}} & 1  & \multicolumn{1}{c|}{\textbf{58.62}} & \multicolumn{1}{c|}{60.75}          & \multicolumn{1}{c|}{60.08}          & \multicolumn{1}{c|}{42.05}          & \multicolumn{1}{c|}{75.72}          & \multicolumn{1}{c|}{72.88}          & \multicolumn{1}{c|}{60.70}          & \multicolumn{1}{c|}{52.66}          & \multicolumn{1}{c|}{65.80}          & \multicolumn{1}{c|}{62.03}          & \multicolumn{1}{c|}{48.43}          & 64.01          & 60.31                 \\ \cline{2-15} 
\multicolumn{1}{c|}{}                                                                                              & 2  & \multicolumn{1}{c|}{59.95}          & \multicolumn{1}{c|}{\textbf{69.05}} & \multicolumn{1}{c|}{62.69}          & \multicolumn{1}{c|}{46.76}          & \multicolumn{1}{c|}{71.59}          & \multicolumn{1}{c|}{77.87}          & \multicolumn{1}{c|}{67.26}          & \multicolumn{1}{c|}{46.41}          & \multicolumn{1}{c|}{69.17}          & \multicolumn{1}{c|}{62.59}          & \multicolumn{1}{c|}{45.51}          & 61.91          & 61.73                 \\ \cline{2-15} 
\multicolumn{1}{c|}{}                                                                                              & 3  & \multicolumn{1}{c|}{59.13}          & \multicolumn{1}{c|}{55.69}          & \multicolumn{1}{c|}{\textbf{61.96}} & \multicolumn{1}{c|}{44.80}          & \multicolumn{1}{c|}{55.34}          & \multicolumn{1}{c|}{72.98}          & \multicolumn{1}{c|}{55.22}          & \multicolumn{1}{c|}{58.77}          & \multicolumn{1}{c|}{61.57}          & \multicolumn{1}{c|}{63.90}          & \multicolumn{1}{c|}{47.76}          & 58.98          & 58.01                 \\ \cline{2-15} 
\multicolumn{1}{c|}{}                                                                                              & 4  & \multicolumn{1}{c|}{56.43}          & \multicolumn{1}{c|}{41.58}          & \multicolumn{1}{c|}{52.77}          & \multicolumn{1}{c|}{\textbf{54.36}} & \multicolumn{1}{c|}{49.77}          & \multicolumn{1}{c|}{63.19}          & \multicolumn{1}{c|}{45.31}          & \multicolumn{1}{c|}{54.00}          & \multicolumn{1}{c|}{51.35}          & \multicolumn{1}{c|}{63.11}          & \multicolumn{1}{c|}{46.22}          & 53.72          & 52.65                 \\ \cline{2-15} 
\multicolumn{1}{c|}{}                                                                                              & 5  & \multicolumn{1}{c|}{66.03}          & \multicolumn{1}{c|}{65.98}          & \multicolumn{1}{c|}{69.52}          & \multicolumn{1}{c|}{56.98}          & \multicolumn{1}{c|}{\textbf{70.62}} & \multicolumn{1}{c|}{76.30}          & \multicolumn{1}{c|}{63.08}          & \multicolumn{1}{c|}{68.32}          & \multicolumn{1}{c|}{67.37}          & \multicolumn{1}{c|}{71.14}          & \multicolumn{1}{c|}{53.12}          & 58.43          & \textbf{65.57}        \\ \cline{2-15} 
\multicolumn{1}{c|}{}                                                                                              & 6  & \multicolumn{1}{c|}{64.19}          & \multicolumn{1}{c|}{58.75}          & \multicolumn{1}{c|}{63.27}          & \multicolumn{1}{c|}{48.93}          & \multicolumn{1}{c|}{62.96}          & \multicolumn{1}{c|}{\textbf{75.88}} & \multicolumn{1}{c|}{61.79}          & \multicolumn{1}{c|}{56.18}          & \multicolumn{1}{c|}{60.33}          & \multicolumn{1}{c|}{66.47}          & \multicolumn{1}{c|}{50.45}          & 54.97          & 60.35                 \\ \cline{2-15} 
\multicolumn{1}{c|}{}                                                                                              & 7  & \multicolumn{1}{c|}{53.29}          & \multicolumn{1}{c|}{59.14}          & \multicolumn{1}{c|}{53.89}          & \multicolumn{1}{c|}{45.32}          & \multicolumn{1}{c|}{64.58}          & \multicolumn{1}{c|}{71.26}          & \multicolumn{1}{c|}{\textbf{65.47}} & \multicolumn{1}{c|}{45.70}          & \multicolumn{1}{c|}{60.99}          & \multicolumn{1}{c|}{65.09}          & \multicolumn{1}{c|}{49.98}          & 58.49          & 57.77                 \\ \cline{2-15} 
\multicolumn{1}{c|}{}                                                                                              & 8  & \multicolumn{1}{c|}{52.23}          & \multicolumn{1}{c|}{45.31}          & \multicolumn{1}{c|}{52.86}          & \multicolumn{1}{c|}{52.30}          & \multicolumn{1}{c|}{56.82}          & \multicolumn{1}{c|}{60.65}          & \multicolumn{1}{c|}{54.31}          & \multicolumn{1}{c|}{\textbf{61.59}} & \multicolumn{1}{c|}{58.43}          & \multicolumn{1}{c|}{60.64}          & \multicolumn{1}{c|}{51.43}          & 49.09          & 54.64                 \\ \cline{2-15} 
\multicolumn{1}{c|}{}                                                                                              & 9  & \multicolumn{1}{c|}{60.64}          & \multicolumn{1}{c|}{72.23}          & \multicolumn{1}{c|}{62.85}          & \multicolumn{1}{c|}{49.89}          & \multicolumn{1}{c|}{58.22}          & \multicolumn{1}{c|}{76.83}          & \multicolumn{1}{c|}{61.03}          & \multicolumn{1}{c|}{57.43}          & \multicolumn{1}{c|}{\textbf{77.62}} & \multicolumn{1}{c|}{71.42}          & \multicolumn{1}{c|}{59.46}          & 60.94          & 64.05                 \\ \cline{2-15} 
\multicolumn{1}{c|}{}                                                                                              & 10 & \multicolumn{1}{c|}{59.67}          & \multicolumn{1}{c|}{62.55}          & \multicolumn{1}{c|}{64.04}          & \multicolumn{1}{c|}{57.62}          & \multicolumn{1}{c|}{58.70}          & \multicolumn{1}{c|}{75.54}          & \multicolumn{1}{c|}{70.99}          & \multicolumn{1}{c|}{50.39}          & \multicolumn{1}{c|}{66.17}          & \multicolumn{1}{c|}{\textbf{74.15}} & \multicolumn{1}{c|}{54.27}          & 58.55          & 62.72                 \\ \cline{2-15} 
\multicolumn{1}{c|}{}                                                                                              & 11 & \multicolumn{1}{c|}{52.03}          & \multicolumn{1}{c|}{30.06}          & \multicolumn{1}{c|}{47.69}          & \multicolumn{1}{c|}{30.53}          & \multicolumn{1}{c|}{41.49}          & \multicolumn{1}{c|}{60.51}          & \multicolumn{1}{c|}{49.30}          & \multicolumn{1}{c|}{39.93}          & \multicolumn{1}{c|}{46.64}          & \multicolumn{1}{c|}{46.57}          & \multicolumn{1}{c|}{\textbf{42.41}} & 47.26          & \textbf{44.54}        \\ \cline{2-15} 
\multicolumn{1}{c|}{}                                                                                              & 12 & \multicolumn{1}{c|}{54.90}          & \multicolumn{1}{c|}{56.75}          & \multicolumn{1}{c|}{55.42}          & \multicolumn{1}{c|}{46.45}          & \multicolumn{1}{c|}{55.56}          & \multicolumn{1}{c|}{70.37}          & \multicolumn{1}{c|}{50.05}          & \multicolumn{1}{c|}{42.89}          & \multicolumn{1}{c|}{55.03}          & \multicolumn{1}{c|}{60.06}          & \multicolumn{1}{c|}{42.19}          & \textbf{64.25} & 54.49                 \\ \hline
\multicolumn{2}{c|}{Avg.}                                                                                               & \multicolumn{1}{c|}{58.09}          & \multicolumn{1}{c|}{56.49}          & \multicolumn{1}{c|}{58.92}          & \multicolumn{1}{c|}{48.00}          & \multicolumn{1}{c|}{60.11}          & \multicolumn{1}{c|}{\textbf{71.19}} & \multicolumn{1}{c|}{58.71}          & \multicolumn{1}{c|}{52.86}          & \multicolumn{1}{c|}{61.70}          & \multicolumn{1}{c|}{63.93}          & \multicolumn{1}{c|}{\textbf{49.27}} & 57.55          \\        
\hline
\end{tabular}
\setlength{\belowcaptionskip}{-5pt}
 \vspace{-2mm}
 \caption{Cross-sector test with models trained on the same size of each sector (AUPRC, \%).}
 \vspace{-3mm}
    \label{table:samesector-crosstest}
    
\end{table*}

\begin{table}[]
\footnotesize
    \centering
    
     \begin{tabular}{c|c}
    \hline
    Model & 
    \multicolumn{1}{|c}{AUPRC (\%)} \\
    \hline\hline
    LSTM        & 83.65  \\ 
    Bi-LSTM     & 83.38  \\ 
    ELECTRA Base & 84.92 \\ 
    BERT Base & \textbf{85.13} \\ 
    \hline
    \end{tabular}
    \setlength{\belowcaptionskip}{-12pt}
        \caption{Classification performance of baseline models; LSTM, Bi-LSTM, ELECTRA Base, and BERT Base.}
    
    \label{table:baselines}
\end{table}

\subsection{Experimental Settings}
\label{sec:experimental_settings}
We conduct experiments with baseline models on CR-COPEC consisting of sentences and corresponding labels in supervised learning. 
Since the dataset is highly imbalanced, we use area under the precision-recall curve (AUPRC) of causal sentences as the evaluation metric \citep{davis2006relationship}. Models are optimized for the validation dataset. A trained model gives us the probability of causality on each sentence. 
Experiments are conducted on Google Colab Pro with one Tesla V100-SXM2-16GB GPU and four Intel(R) Xeon(R) CPU @ 2.00GHz CPUs.

\subsection{Experimental Results on Baseline Models}
We use LSTM, Bidirectional LSTM (Bi-LSTM) \cite{graves2005bidirectional}, ELECTRA Base \cite{clark2020electra} and BERT Base as baselines to compare the performance of extracting rationale of financial performance changes. As an input, each sentence is tokenized by ELECTRA and BERT \cite{DBLP:journals/corr/abs-1810-04805} tokenizers respectively for ELECTRA Base and BERT Base. For the LSTM models, we use Glove \cite{pennington2014glove} word embedding. 

For the baseline experiments, we use total CR-COPEC dataset describribed in Section~\ref{sec:Protocols for Total Dataset}. As shown in Table \ref{table:baselines}, models based on transformers (ELECTRA Base and BERT Base) performed better than RNN based models. Because BERT Base achieved the highest AUPRC score (85.13\%), we utilize BERT Base for followed experiments.

    




\begin{figure*}[!ht]
\centering
\begin{center}
\includegraphics[width=.9\linewidth]{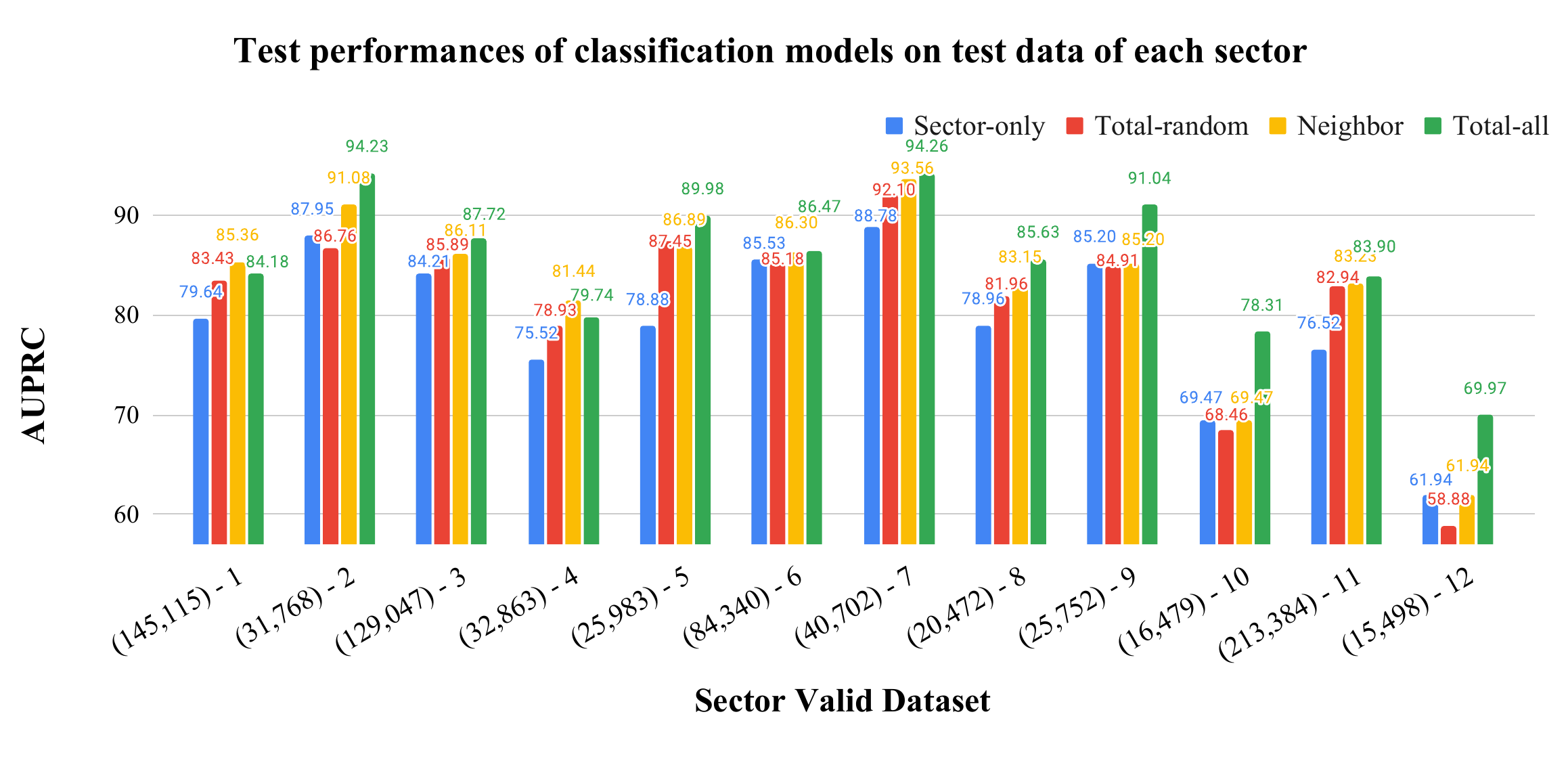}
\end{center}
\setlength{\abovecaptionskip}{-5pt}
\vspace*{-4mm}
\caption{Test performances of classification models on valid dataset of each sector (AUPRC). The size of dataset for Neighbor and Total-random is presented in parenthesis.}
\setlength{\belowcaptionskip}{-14pt}
\vspace*{-5mm}
\label{fig:neighbor}
\end{figure*}

\subsection{Examination of Sectors' Characteristics} 
\label{sec:sectorspecific}
CR-COPEC consists of twelve different industries. We observe that each industry possesses its own field of interests with respect to reasons of financial performance changes. Therefore, training each sector individually is required for a better classification. However, the number of data for each industry varies around [5k-60k], in which some sectors do not contain enough data to detect causal rationale precisely. Thus, this section tests a hypothesis if training other sectors together helps detecting causal sentences in a target sector.
For this purpose, we conduct cross-sector test and compare the performance of models trained on various versions of the CR-COPEC dataset.

First, we fine-tune BERT Base on the equal size of fraction dataset described in Section~\ref{sec:Protocols for Fraction of Dataset} and test across sectors. We call this model as Sector-only Same-size. Then, we select neighbor sectors based on the results from the cross-sector test with valid dataset (Table~\ref{table:samesector-crosstest}). If the models trained on a neighbor sector detect target sector sentences better than the model trained on target sector, we assume that this sector can support to train a model. For example, in the case of Sector 8, we select Sector 5 as a neighbor which shows 68.32\% in AUPRC ($> 61.59\%$). 
After we select neighbors for each sector, we fine-tune BERT Base with the target and all selected neighbor sector datasets. We define this model as Neighbor model. Besides, for equal size comparison, we randomly select data from Total dataset to the same size of each Neighbor model trained on. The model trained on this dataset is called Total-random. In addition, we also train model on the whole size of each sector dataset and call it as Sector-only. Finally, we name BERT Base trained on Total dataset from the Section~\ref{sec:experimental_settings} as Total-all. We compare the performances of these four different types of models on in details below.

From the results in Table~\ref{table:samesector-crosstest}, we also find that the average performance of models on Sector 6 is the highest (71.19\%) and Sector 11  (49.27\%) is the lowest. This result demonstrates a gap in the level of difficulty of detecting causality on target sectors. Detecting causal sentences in Sector 6 is relatively easier and on Sector 11 is more difficult than other sectors. Furthermore, Sector-only Same-size model trained on fraction dataset of Sector 5 shows the highest average performances across all sectors. Meanwhile, the model trained on fraction dataset of Sector 11 shows the lowest average score. We interpret these results as the sector generality and specificity. It can be considered as Sector 5 consists of more general causality patterns that can support to other sectors easier, while Sector 11 contains sector specific information so harder to transferable to other sectors. Hence, we find that Sector 11 is difficult to find causal sentences and also hard to be transferable to other sectors. We assume that since Sector 11 is the sector of finance industry, its unique characteristics present the sector specificity in our dataset. However, this transferable interaction between sectors is asymmetric, so other sectors can easily be a help to Sector 11.

Figure~\ref{fig:neighbor} compares the performances of Sector-only, Total-random, Neighbor and Total-all models on each sector. Overall, Neighbor models show significant higher performances compared to Sector-only and Total-random. We conduct t-tests of Neighbor and Total-random to determine the statistical significance of the difference in performances of AUPRC ($p < .01$). In Sector 1 and 4, Neighbor is higher than Total-all, though the training dataset's size is much smaller that Total dataset.

\subsection{Effects of Dataset Sizes}
\label{sec:num_of_train_data}

\begin{figure}[t]
\centering
\includegraphics[width=\linewidth]{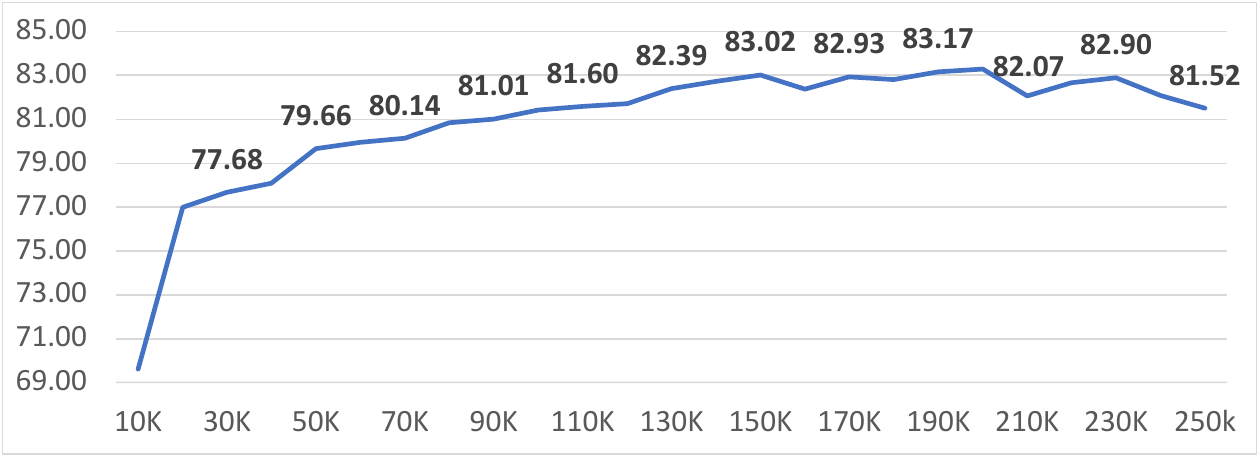}
\caption{AUPRC of BERT Base by the number of training data from CR-COPEC.}
\setlength{\belowcaptionskip}{-20pt}
\vspace*{-4mm}
\label{fig:numberofdata}
\end{figure}

We check the effect of the number of training data with BERT Base trained on CR-COPEC. We increase the amount of training data and verify the causal extracting performance on each step. At each step, we randomly select a set of training data by increasing 10,000 sentences, then report AUPRC on CR-COPEC test dataset.
AUPRC of BERT Base increased with a huge gap (6.64\%) from 10K to 20K. Then they increase gradually until 200K then slightly drop afterward. We provide the detailed results in Figure~\ref{fig:numberofdata}. Figure~\ref{fig:neighbor} shows that the gap between Total-all and Neighbor is bigger when the number of training data is relatively smaller than Total dataset, in case of Sector 9, 10 and 12.

\begin{figure*}[!ht]
\centering
\begin{center}
\includegraphics[width=.80\linewidth]{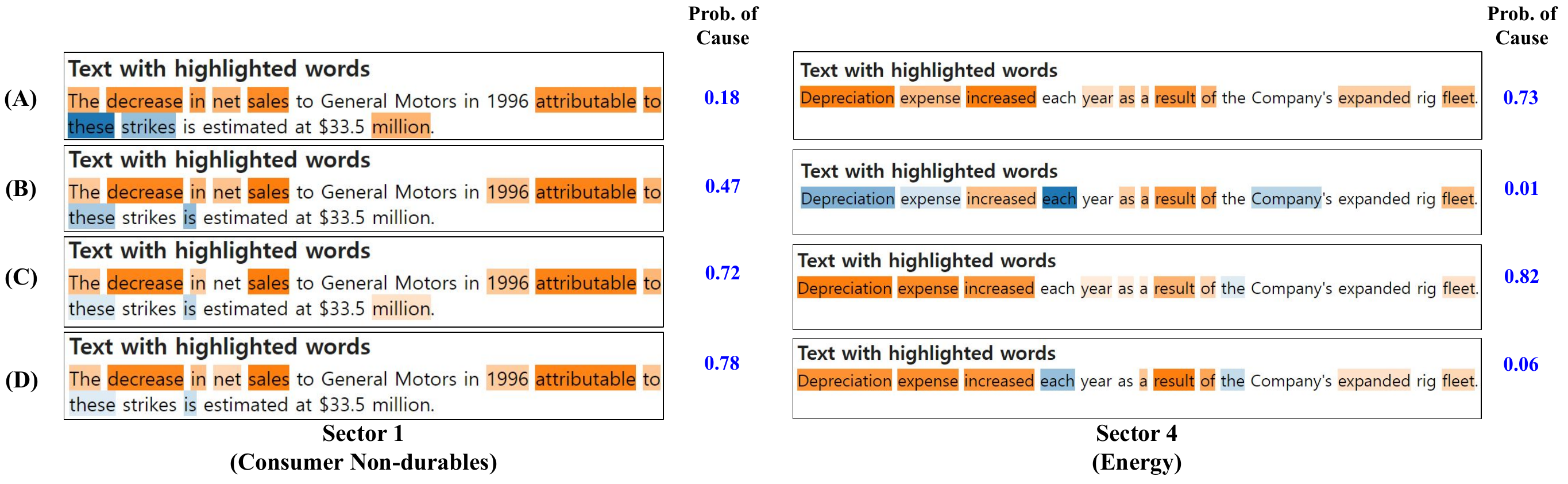}
\end{center}
\setlength{\abovecaptionskip}{-2pt}
\caption{Interpretation of predictions from (A)`Sector-only',  (B)`Total-random', (C)`Neighbor' and (D)`Total-all' models with LIME. Features contributing on causal rationale are highlighted in orange, the opposites are in blue.}
\setlength{\belowcaptionskip}{-14pt}
\vspace*{-5mm}
\label{fig:lime}
\end{figure*}

\subsection{Explaining Detected Causal Rationales} 
\label{sec:interpretation_of_causal_rationale_detection}
We analyze input features used for causal rationale classification. Local Interpretable Model-agnostic Explanations (LIME) \cite{lime} is an explanation method that makes the predictions of a classifier interpretable by learning a linear interpretable model locally around the prediction. With this explanation technique, we can visualize the most important features affecting the prediction. To see the difference on the predicted sentences from various models, we applied LIME on our trained models (Figure~\ref{fig:lime}). 

The probability of a causal sentence from Sector 1 is predicted as highest with Total-all, compared to Sector-only, Total-random and Total-all models. The Neighbor model also accurately predicts the correct answer ($> 0.5$). Furthermore, the word `strike' works as negative to causal rationale in Sector-only model. However, it becomes neutral in models trained on more data.

In the example of Sector 4, sector specific models including Sector-only and Neighbor correctly predict the sentence as causal. Meanwhile, Total-random model regard the term `depreciation expense' as negative one at prediction. This is because sentences containing causal rationale of `depreciation expense' are annotated differently across sectors, so training all sectors together brings confusion in this case.

\subsection{Comparisons with an Related Dataset}
\label{sec:cross_dataset}
In order to verify the novelty of CR-COPEC, we compare our data with the existing financial causality detection corpus.
For this, we conduct cross-dataset experiments between CR-COPEC and the FinCausal task 1 \citep{mariko2020financial}.
In specific, we compare the test performances in a cross-dataset setting where we fine-tune BERT Base on 1) FinCausal practice dataset, then test on both our test dataset and FinCausal trial dataset, and 2) CR-COPEC train dataset, then test on both of test dataset and FinCausal trial dataset.

Table~\ref{tab:cross_dataset} show that the performance of each model trained on a specific dataset is obviously decreased in its counterpart. This indicates that there are significant differences between the two datasets and the existing dataset cannot fully address the causality detection in 10-K reports and vice versa. We believe this is owing to the purpose and the target text of the datasets being different. Specifically, the CR-COPEC dataset aims to extract causal rationale sentences based on accounting items that are able to cause companies' performance changes. Besides, CR-COPEC targets formal public text. On the other hand, FinCausal targets relatively casual text since FinCausal dataset tries to elicit comprehensive causality texts in financial documents.

\begin{table}[]
\footnotesize
\centering
\begin{tabular}{c|c|c}
\hline
\diagbox[innerwidth = 3cm, height = 2.75ex]{Train}{Test}& CR-COPEC & FinCausal \\
\hline
\hline
CR-COPEC  & 85.13 &\textbf{12.81}\\
FinCausal &  \textbf{9.75}& 69.94\\ 
\hline
\end{tabular}

\setlength{\belowcaptionskip}{-7pt}
\vspace{-2mm}
\caption{The experimental results of the cross-dataset between CR-COPEC and FinCausal dataset (AUPRC).}
\vspace{-2mm}
\label{tab:cross_dataset}
\end{table}

\begin{table}[]
    \centering
    \footnotesize
    \begin{tabular}{c|cccc}
    \hline
        & 1 & 2 & 3 \\
    \hline\hline
        Longformer (AUPRC) & 84.50 & 84.44 & 84.85\\
    \hline
    \end{tabular}
    \vspace{-2mm}
    \caption{The performances (AUPRC) on the difference size of input sentence n-grams ($n=1,2,3$). }
    \vspace{-6mm}
    \label{tab:multi_sentence_modeling}
\end{table}

\subsection{Multiple Sentence Modeling}
\label{sec:multiple_sentence_modeling}
Since the evidence for a causal statement can be scattered across more than one sentence, we conducted additional experiments on sentence n-gram settings to provide context for causal sentences. Each sentence n-gram consists of one target sentence and n-1 previous sentences, which serve as the context. Moreover, by following \citet{mariko2020financial}'s setting, we conducted experiments for cases where $n=1,2,3$, that the length of the context can have up to two sentences. Finally, since the number of tokens in sentence n-grams often exceeds the maximum token input size of BERT (512 tokens), all experiments were performed with Longformer Base \citep{beltagy2020longformer}, which accepts longer inputs (4,096 tokens). Appendix~\ref{apx:multi_sentence_modeling} provides detailed descriptions of the experimental settings. Experimental results of Table~\ref{tab:multi_sentence_modeling} shows that the performance of the multi-sentence (bi- and tri-gram) modelings achieves slightly higher performance at tri-gram modeling setting than single the uni-gram modeling but it is not significant ($p>0.2$). The result implies that, in many cases, the evidences for a causal statement can be found within one sentence.

\subsection{Experimental Results on a Large Language Model}
\label{sec:chatgpt}

\begin{table}[]
\footnotesize
    \centering
    
     \begin{tabular}{l|c|c}
    \hline
    Model & Voting&
    \multicolumn{1}{|c}{F1 Score} \\
    \hline\hline
    BERT Base & - & 0.870 \\
    \hline
    GPT-3.5-Turbo Zero Shot& 1 & 0.698 \\
    GPT-3.5-Turbo Zero Shot& 5 & 0.714 \\
    \hline
    GPT-3.5-Turbo 5-Shot & 1 &  0.734 \\
    
    GPT-3.5-Turbo 5-Shot & 5 & 0.747 \\
    \hline
    \end{tabular}
    \setlength{\belowcaptionskip}{-12pt}
        \caption{\revised{Classification performance of 1,000 sampled test set on our baseline and GPT-3.5-Turbo. Voting indicates the number of voters participating in inference based on majority voting. When voting is 1, it means single inference. Otherwise, when voting is 5, it means the majority voting of 5 times inferences.}}
    
    \label{table:chatgpt}
\end{table}
\revised{This section reports the experimental results on a state-of-the-art large language model (ChatGPT; GPT-3.5-Turbo) \citep{ouyang2022training} with our baseline model (BERT Base). For this, 1,000 instances were randomly sampled from the test set. The experiments were performed in the prompting-based zero-shot inference (Zero Shot) and 5-shot inference \citep{brown2020language} where 5 positive and negative examples from the same sector as the input text were randomly selected from the training set. In addition, we conducted experiments on both of single inference and majority voting five inferences by following \citet{arora2022ask}'s setting. Detailed experimental settings and prompts are described in Appendix~\ref{apx:chatgpt}.}

\revised{Table~\ref{table:chatgpt} shows that ChatGPT can successfully carry out a specific part of causal rationale text analysis without additional fine-tuning or zero-shot in-context learning. Additionally, 5-shot settings outperform zero-shot settings approximately 3\%p. Furthermore, it has been demonstrated that employing majority voting surpasses the single inference-based performance. However, those performances are still significantly lower ($p<10^{-4}$) than that of the fine-tuned smaller model; BERT Base. This is because ChatGPT's results are sensitivity to minor fluctuations in financial condition or change in minor accounting items within (Interest income in the Sector 1) the company's industrial sector. These results imply the need for an annotated corpus based on the domain knowledge on the specialized field even in the large language model's era.}


\subsection{Potential Uses of Dataset}
Financial analysts are key information intermediaries in capital markets, with their research, focused on uncovering private information and interpreting public data, being highly valued by investors. The value of analyst research can stem from two main sources: uncovering private information and interpreting public information, as exemplified by studies such as \citep{ivkovic2004timing, asquith2005information}. To uncover this private information and reinterpret public information, analysts often analyze various linguistic patterns in annual reports. Given the sheer volume of corporate filings today, not only are they paying significant time and money to analyze these reports, but their relying on manual annotation also reduces the accuracy of the analysis. Otherwise, NLP models trained with our dataset can automates annotations enhancing both speed and scalability.
Moreover, analysts typically specialize in specific areas, requiring domain knowledge in industry specifics and market patterns. This includes understanding industry dynamics, competitive landscape, and regulatory environment, along with learning unique financial metrics and varying accounting practices across industries. Analysts may need to grasp new metrics and varying accounting practices across industries to understand how financial statements are prepared and analyzed, allowing them to make meaningful comparisons. Our dataset is expected to significantly reduce the learning curve and associated costs for analysts transitioning between industries.

\section{Conclusion}
We introduce a novel large scale dataset to extract causal rationales from financial reports in various sector. 
The dataset was annotated by considering accounting factors according to the industry.
We further validate the process of building our CR-COPEC dataset. Finally, through the qualitative and quantitative analysis, we observed that the model trained with CR-COPEC recognize the clue of causal sentences.
We hope that our work would promote the study of causality detection in the financial text domain.

\section*{Acknowledgement}
Institute of Information \& communications Technology Planning \& Evaluation (IITP) grant funded by the Korea government (MSIT) (No.2022-0-00984, Development of Artificial Intelligence Technology for Personalized Plug-and-Play Explanation and Verification of Explanation, No.2019-0-00075, Artificial Intelligence Graduate School Program (KAIST), NO.2022-0-00184, Development and Study of AI Technologies to Inexpensively Conform to Evolving Policy on Ethics).

\section*{Ethical Consideration}
The main ethical concern in this work is a data license and releasing issue of 10-K database. First of all, 10-K database is publicly available, so a license is not required to access the database. In addition, according to information presented on \url{www.sec.gov} is considered public information and may be copied or further distributed by users of the web site without the SEC’s permission. Secondly, to all annotators, we paid them by obeying the minimum wage standard of national law. Besides, the general annotators were full-time employees.

\section*{Limitation}
CR-COPEC may include two types of bias: 1) protocol bias, 2) annotators' subjectivity bias. First of all, we clearly defined in this paper that we annotate sentences based on annotation rules regarding financial statement items. Thus, our dataset may have different criteria from other causality detection problems (e.g, FinCausal dataset). Secondly, we are aware that the subjective judgment of annotators may exist in the annotation process. Therefore, inconsistency of annotations between annotators may occur for some sentences, which may affect data quality degradation and model performance. To control this issue, we verified the human annotation process (Section~\ref{sec:veryfying_annotation}) and obtained substantial agreement results by measuring the kappa score between general annotators and additional annotators for 1000 sample sentences. The reliability of each general annotator is 0.7 or higher, and the agreement between general annotators is 0.698. 

CR-COPEC is designed to detect causal statements that can limit further analyses. In the future, those  without considering the positive or negative impact of performance changes. Additional analysis based on subcategories such as positive/negative/neutral on the impact of causal statements on a corporate performance changes is required in the future.

\bibliographystyle{acl_natbib}
\bibliography{anthology,emnlp2021}

\onecolumn
\appendix
\section{Annotation Guideline}
\label{sec:appendix_annotation_guideline}
\begin{table*}[ht!]
\centering
\small
\rotatebox{0}{
\begin{tabular}{c@{ }|@{ }c@{ }|l@{ }|@{ }cccccccccccc}
\hline
\multirow{2}{*}{Main Category} & \multirow{2}{*}{Sub-category} & \multicolumn{1}{c@{ }|@{ }}{\multirow{2}{*}{Items}} & \multicolumn{12}{c}{Sector} \\
\cline{4-15}
 & & \multicolumn{1}{@{ }c@{ }|@{ }}{} & \multicolumn{1}{l}{1} & \multicolumn{1}{l}{2} & \multicolumn{1}{l}{3} & \multicolumn{1}{l}{4} & \multicolumn{1}{l}{5} & \multicolumn{1}{l}{6} & \multicolumn{1}{l}{7} & \multicolumn{1}{l}{8} & \multicolumn{1}{l}{9} & \multicolumn{1}{l}{10} & \multicolumn{1}{l}{11} & \multicolumn{1}{l}{12} \\
 \hline
\multirow{26}{*}{\begin{tabular}[c]{@{}c@{}}Income\\ Statements\end{tabular}} & \multirow{6}{*}{Sales} & Operating revenue & 1 & 1 & 1 & 1 & 1 & 1 & 1 & 1 & 1 & 1 & 1 & 1 \\
\cline{3-15}
 & & Service revenue & 0 & 0 & 0 & 0 & 0 & 0 & 0 & 0 & 1 & 0 & 0 & 1 \\
 \cline{3-15}
 & & Engineering service revenue & 0 & 1 & 1 & 0 & 0 & 1 & 0 & 0 & 0 & 0 & 0 & 0 \\
 \cline{3-15}
 & & Oil and gas oil and gas revenue & 0 & 0 & 0 & 1 & 0 & 0 & 0 & 0 & 0 & 0 & 0 & 0 \\
 \cline{3-15}
 & & Lease revenue & 1 & 0 & 0 & 0 & 0 & 0 & 0 & 0 & 0 & 0 & 0 & 0 \\
 \cline{3-15}
 & & Contract logistics revenue & 0 & 0 & 0 & 1 & 0 & 0 & 0 & 0 & 0 & 0 & 0 & 0 \\
 \cline{2-15}
 & Cost & Sale cost & 1 & 1 & 1 & 1 & 1 & 1 & 1 & 1 & 1 & 1 & 1 & 1 \\
 \cline{2-15}
 & \multirow{3}{*}{Gross Profit} & Net sales & 1 & 1 & 1 & 1 & 1 & 1 & 1 & 1 & 1 & 1 & 1 & 1 \\
 \cline{3-15}
 & & Gross profit & 1 & 1 & 1 & 1 & 1 & 1 & 1 & 1 & 1 & 1 & 1 & 1 \\
 \cline{3-15}
 & & Gross profit margin & 1 & 1 & 1 & 1 & 1 & 1 & 1 & 1 & 1 & 1 & 1 & 1 \\
 \cline{2-15}
 & \multirow{6}{*}{\begin{tabular}[c]{@{}c@{}}Operating\\Expenses\end{tabular}} & Research development expense & 0 & 0 & 0 & 0 & 1 & 1 & 0 & 0 & 0 & 1 & 0 & 0 \\
 \cline{3-15}
 & & Amortization expense & 0 & 0 & 0 & 0 & 1 & 1 & 0 & 0 & 0 & 1 & 0 & 0 \\
 \cline{3-15}
 & & Depreciation expense & 0 & 0 & 1 & 1 & 0 & 0 & 0 & 0 & 0 & 0 & 0 & 1 \\
 \cline{3-15}
 & & Legal and entitlement cost & 0 & 0 & 0 & 0 & 0 & 0 & 0 & 0 & 0 & 0 & 1 & 0 \\
 \cline{3-15}
 & & \begin{tabular}[c]{@{}l@{}}Expenditure associated with \\ Environmental liabilities\end{tabular} & 0 & 0 & 0 & 0 & 0 & 0 & 0 & 0 & 0 & 0 & 0 & 1 \\
 \cline{2-15}
 & \begin{tabular}[c]{@{}c@{}}Operating\\Incomes\end{tabular} & Operating income & 1 & 1 & 1 & 1 & 1 & 1 & 1 & 1 & 1 & 1 & 1 & 1 \\
 \cline{2-15}
 & \multirow{6}{*}{\begin{tabular}[c]{@{}c@{}}Non-operating\\Items\end{tabular}} & Non operating income & 1 & 1 & 1 & 1 & 1 & 1 & 1 & 1 & 1 & 1 & 1 & 1 \\
 \cline{3-15}
 & & (net) Interest expense & 0 & 0 & 0 & 0 & 0 & 0 & 0 & 0 & 0 & 0 & 1 & 0 \\
 \cline{3-15}
 & & Interest income & 0 & 0 & 0 & 0 & 0 & 0 & 0 & 0 & 0 & 0 & 1 & 0 \\
 \cline{3-15}
 & & \begin{tabular}[c]{@{}l@{}}Purchased services and \\ other expenses\end{tabular} & 0 & 0 & 0 & 0 & 0 & 0 & 0 & 0 & 0 & 0 & 0 & 1 \\
 \cline{3-15}
 & & Loan payment & 0 & 1 & 1 & 0 & 0 & 0 & 0 & 0 & 0 & 0 & 0 & 0 \\
 \cline{3-15}
 \cline{2-15}
 & Net Income & Net income & 1 & 1 & 1 & 1 & 1 & 1 & 1 & 1 & 1 & 1 & 1 & 1 \\
 \cline{2-15}
 & - & Manufacturer price & 1 & 1 & 1 & 1 & 1 & 1 & 1 & 1 & 1 & 1 & 1 & 1 \\
\hline
\multirow{2}{*}{\begin{tabular}[c]{@{}c@{}}Balance\\ Sheets\end{tabular}} & Liabilities & Debt & 0 & 0 & 0 & 0 & 1 & 1 & 0 & 0 & 0 & 1 & 0 & 0 \\
\cline{2-15}
 & Assets & Mortgage loan & 0 & 0 & 0 & 0 & 0 & 0 & 0 & 0 & 0 & 0 & 1 & 0 \\
 \hline
\end{tabular}}
\caption{The annotation guideline for each sector. Herein, 1 and 0 indicate signals for \textit{causal rationale of sentences} and \textit{non-causal rationale of sentences} respectively. Note that \textbf{the guideline is not an absolute standard}, and it is \textbf{possible to flexibly annotate} if it is judged as causal sentences according to the annotator's point of view. For items not included in this guideline, the causality of a target sentence is annotated by depending on annotators' subjection. }
\end{table*}

\section{Main Factors}
\label{sec:main_factors}
\begin{figure}[!ht]
\begin{center}
\includegraphics[width=.9\linewidth]{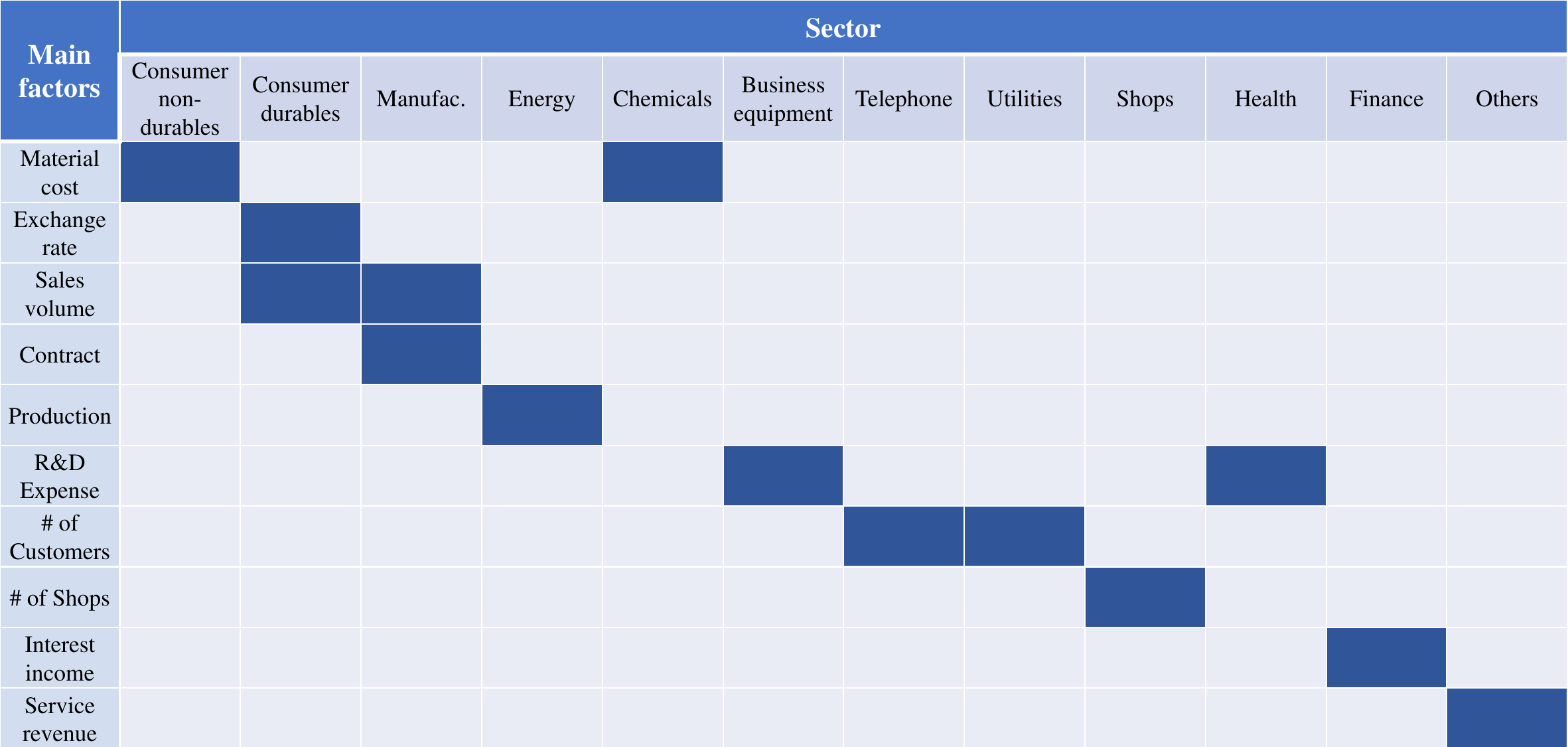}
\end{center}
\caption{Examples of each sector's main factors.}
\label{fig:main_factors}
\end{figure}

\newpage
\section{Additional Examples of Causal and Non-causal rationale of sentences}
\label{sec:appendix_additional_examples_of_causal_non_causal_sentences}


\label{sec:supplemental}

\begin{table*}[!h]
    \centering
    \small
    \begin{tabular}{p{0.97\columnwidth}}
    \hline
    \textbf{Sector of Industry} - 
    {{Example}} {{[Document]}} \\
    \hline
    \hline
    \textbf{Consumer Non-Durables} - The gross profit margin as a percentage of sales improved from 44.3\% in Fiscal Year 1995 to 46.8\% in Fiscal Year 1996, principally due to lower green coffee and material costs, and lower plant overhead costs. [Brothers Gourmet Coffees, Inc., July, 1997] \\
    \hline
    \textbf{Consumer Durables} 
    - Other income (expense), net increased primarily as a result of foreign currency exchange gains. [Breed Technologies, Inc., September, 1997]\\
    \hline
    \textbf{Manufacturing} -
    The increase in 1996 net sales was due primarily to increases in sales revenues recognized on the contracts to construct the first five Sealift ships, the Icebreaker and the forebodies for four double-hulled product tankers, which collectively accounted for 63\% of the Company's 1996 net sales revenue. [Avondale Industries, Inc., March, 1997]\\
    \hline
    \textbf{Energy} - Oil and gas sales decreased due to a decrease in production volumes. [Castle Energy Corp., December, 1997]\\
    \hline
    \textbf{Chemicals} - Revenues decreased eight percent to \$3,955 million in 1996 from \$4,282 million in 1995 primarily reflecting lower SM prices and, to a lesser extent, lower volumes for PO and derivatives.[Arco Chemical Co., February, 1997]\\
    \hline
    \textbf{Business Equipment} - 
    Research and development expenses increased from \$10.1 million (2\% of net revenues) in fiscal 1995 to \$34.6 million (21\% of net revenues) in fiscal 1996 due to the increase in Software development resulting from the acquisition of the three Software studios in calendar 1995. [Acclaim Entertainment, Inc., November, 1997]\\
    \hline
    \textbf{Telephone} - 
    Revenue from cable television operations increased by \$90,713 or 24.6\%, over the corresponding year ended May 31, 1996 as a result of regulated price increases, increases in the number of cable television subscribers and acquisitions. [Century Communications Corp., August, 1997]\\
    \hline
    \textbf{Utilities} - 
    Gas operating revenues increased \$36.7 million, or 21.0\%, due to increased volumes as a result of customer growth and higher gas costs. [WPS Resources Corp., March, 1997]\\
    \hline
    \textbf{Shops} - 
    Aggregate sales generated by franchised stores increased by \$10,001,000, or 13.1\%, to \$86,485,000 for calendar year ended December 31, 1995, as compared to \$76,484,000 for the same period in 1994, due to an increase of the number of franchised stores, as well as higher sales volume per store. [Sterling Vision, Inc., April, 1997]\\
    \hline
    \textbf{Health} - 
    The increase in research and development expenses in 1996 and 1995 was due primarily to higher expenditures for the Actiq Cancer Pain Program, new product development and other expenditures for product development, including clinical trials. [Anesta Corp., March, 1997]\\
    \hline
    \textbf{Finance} - 
    The investment income decrease resulted from a declining asset base, in large part resulting from loan repayments.\\
    \hline
    \textbf{Others} - 
    The Company's largest revenue source is from the marketing and administration of extended vehicle service contracts (``VSCs") under the EasyCare(R) name, which provided 99\% of revenues for 1996. [Automobile Protection Corp-AUPRCCO, March, 1997]\\
    \hline
    \end{tabular}
        \caption{Additional examples of causal rationale of sentences by each sector.}
        \label{table:add_cause}
\end{table*}

\begin{table*}[!h]
    \centering
    \small
    \begin{tabular}{p{0.97\columnwidth}}
    \hline
    \textbf{Section of Industry} - 
    {{Example}} {{[Document]}} \\
    \hline
    \hline
    \textbf{Consumer Non-Durables} - Total general and administrative costs decreased by \$79,000 in 1995 due primarily to the absence of a management fee for 1995. [Highwater Ethanol, LLC, January, 2017]\\
    \hline
    \textbf{Consumer Durables} - Bank borrowings during 1995 were attributable to the Silver Furniture acquisition and the refinancing of Silver Furniture's bank indebtedness. [Chromcraft Revington, Inc., March, 1997]\\
    \hline
    \textbf{Manufacturing} - Because components are sold directly to the Company's manufacturing sources, the Company is not aware of the precise quantities sourced from particular suppliers.  [Fossil, Inc., March, 1997]\\
    \hline
    \textbf{Energy} - 
    Due to the apparent age of the material, no fine or enforcement action is expected. [Arabian Shield Development Co, March, 1997]\\
    \hline
    \textbf{Chemicals} - Due to personnel additions to the department, employee wages increased approximately \$56,700 in 1996. [American Vanguard Corp., March, 1997]\\
    \hline
    \textbf{Business Equipment} - Government contracts are subject to negotiated overhead rates, and work performed under government contracts is subject to audit and adjustments of amounts paid to the Company. [Ibis Technology Corp., 1997]\\
    \hline
    \textbf{Telephone} - Cost of services related to the wireless telephone operations during the year ended May 31, 1996 was \$26,129, an increase of \$3,977 or 18.0\% as compared to the year ended May 31, 1995. [Century Communications Corp., August, 1997]\\
    \hline
    \textbf{Utilities} - The remainder of the increase was attributable to increases in ad valorem taxes, repair and maintenance expense mainly related to the WCLSF and the employee incentive plan which rewards certain of Tejas' employees with bonuses when the company achieves certain annual financial growth targets. [Tejas Gas Corp., March, 1997]\\
    \hline
    \textbf{Shops} - The Board may increase or decrease the number of shares under the Program or terminate the Program in its discretion at any time. [Boise Cascade Co., February, 2017]\\
    \hline
    \textbf{Health} - The 1995 results were also negatively impacted by a reduction of the Company's income tax benefit resulting from reserves established related to the expiration of certain state operating losses. [American White Cross Inc., April, 1997]\\
    \hline
    \textbf{Finance} - In the last three years, inflation has not had a significant impact on the Company because of the relatively low inflation rate. [Weeks Corp., March, 1997]\\
    \hline
    \textbf{Others} - In addition, the timing of revenue is difficult to forecast because the Company's sales cycle is relatively long. [Claremont Technology Group Inc., September, 1997]\\
    \hline
    \end{tabular}
        \caption{Examples of non-causal rationale of sentences by each section.}
        \label{table:non-cause}
\end{table*}


\section{Details on Annotation Environment}
\label{sec:annotation_env}
\begin{figure}[ht]
    \centering
    \includegraphics[width=\textwidth]{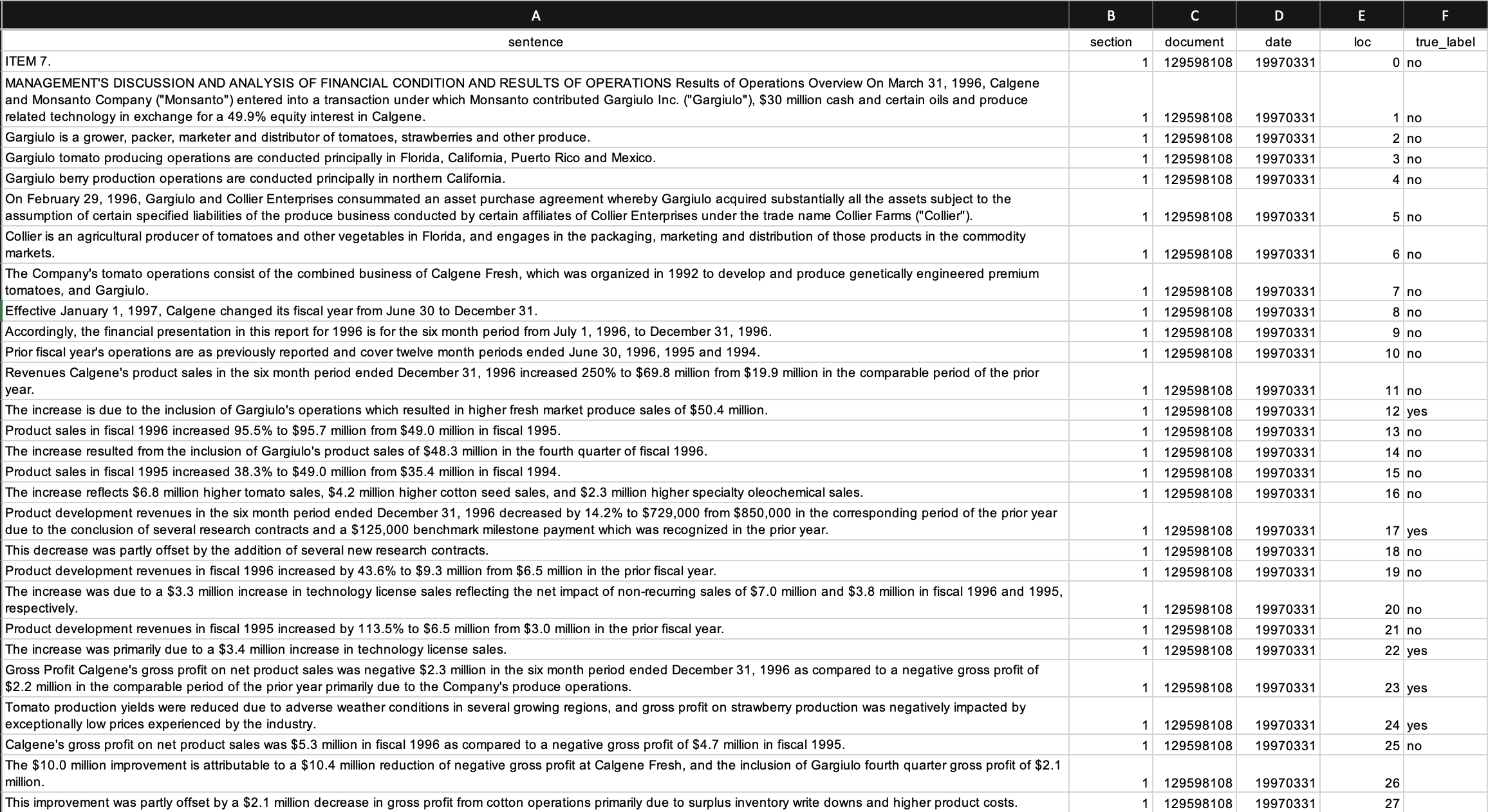}
    \caption{Illustrative example of the annotation form.}
    \label{fig:annotation_env}
\end{figure}

\noindent Annotation files are based on the Comma-Separated Values (CSV) format. In this case, all text in each report of a corporation is gathered in a single CSV file. The file consists of 6 columns: sentence, section, document, date, location (loc), and true$\_$label. The sentence column contains the text information of the report that is split by each sentence. The section column indicates the SIC code of the corporation. The document column represents the unique document ID of the report. Date is the date the report was published. In addition, the location (loc) column signifies the sentence number of the text. Finally, true$\_$label is the human annotated label for a sentence.

Each file was assigned to an annotator. The annotator coded sentences based on the annotation guidelines and the previous context. Specifically, for consecutive sentences related to the same accounting item, only the sentences that provide a reason for the performance change are labeled as causal rationale. For example, in the sentences at location 11, although product sales (which could be a causal rationale for performance change of a company with section 1) increased, it is difficult to consider the sentence as a causal rationale since the reason is not explained. On the other hand, sentences at location 12 are regarded as a causal rationale because the reason for the increment in product sales is specified and clear in the previous context. Note that, since the document and location information are also provided in the final version of the file, it is possible on modeling and inference for multiple sentences as in the experiments in section~\ref{sec:multiple_sentence_modeling}.

\newpage
\section{Comparison Between FinCausal and CR-COPEC}
\label{apx:dataset_comparison}

In this section, we compare CR-COPEC with the most relevant dataset FinCausal \citep{mariko2020financial}, specifically FinCausal task 1. Both of two datasets target to extract causality detection in the financial domain and a binary classification. 

\begin{itemize}
    \item Goal
        \begin{itemize}
            \item FinCausal: If a text contains a causal scheme
            \item CR-COPEC: If a sentence contains a causal rationale of changes in corporate financial performance.
        \end{itemize}
    \item Data Source
        \begin{itemize}
            \item FinCausal: 2019 financial news provided by Qwam
            \item CR-COPEC: 1997 and 2017 10-K reports
        \end{itemize}
    \item Data Size
        \begin{itemize}
            \item FinCausal: 31,580 text section
            \item CR-COPEC: 1,584 10-K reports with 283,490 sentences
        \end{itemize}
    \item Class Bias
        \begin{itemize}
            \item FinCausal: 7.24\%
            \item CR-COPEC: 3.93\%
        \end{itemize}
\end{itemize}

However, there are some fundamental differences between the two datasets that make our data unique. First of all, the Goal of the two datasets are different. FinCausal aims to find \textbf{a causal scheme in a text}, while our dataset targets to identify more specific information that is \textbf{a causal rationale of changes of financial performance}. It indicates that CR-COPEC solely concentrates on causal rationales of \textbf{accounting items considering the unique characteristics of various industries}. Secondly, FinCausal collected their data from financial news. Otherwise CR-COPEC utilizes 10-K reports. It means that \textbf{CR-COPEC is written in a formal manner} because they should comply with regulatory rules. Furthermore, the data size of CR-COPEC is larger than FinCausal. Finally, both of datasets are biased but 
the causal class distribution of the two datasets is different.

\newpage
\section{Details on Multi-Sentence Modeling Experiments}
\label{apx:multi_sentence_modeling}
This section explains in detail experimental settings for the multi-sentence modeling in Section~\ref{sec:multiple_sentence_modeling}. We first briefly introduce the sentence n-gram setting. Then, we describe how we constructed inputs for sentence n-grams. Note that, since sentence n-grams are often longer than the maximum input token size (512 tokens), we utilized Longformer Base, which can handle longer input tokens (4,096 tokens), for the experiments. We used the same hyper-parameter setting of BERT fine-tuning.

In this paper, sentence n-grams consist of 1 target sentence and n-1 previous context sentences. For example, let's suppose the i$^th$ sentence ($s_i$) is a target sentence for which we want to predict whether it is a causal sentence or not. A sentence uni-gram only consists of $s_i$ itself, which is the same as the general single sentence modeling. Otherwise, a sentence bi-gram is made up of $s_i$ and the $s_{i-1}$ context sentence. Similarly, a sentence tri-gram contains $s_i$ and two context sentences, $s_{i-1}$ and $s_{i-2}$.


\begin{center}
    $s_1,~s_2~...~s_{i-2},~s_{i-1},~s_i$
\end{center}

The target sentence and the context are concatenated and input to Longformer. Herein, we mark the boundary between the separate target sentence and context by utilizing Longformer's separator token $<$/s$>$.

\begin{center}
    \textit{\{target sentence place holder\}} $<$/s$>$ \textit{\{context place holder\}}
\end{center}

\section{Details of the Experimental Setting on ChatGPT}
\label{apx:chatgpt}
Herein, since it is difficult to obtain a probability distribution in the prompting-based ChatGPT inference, we set the output as a binary format; `yes' or `no' whether the given input text contains causal rationale. The BERT Base model employs a threshold probability of 0.5. Finally, we set an evaluation metric as the F1 score known to be beneficial for evaluating imbalanced data. Furthermore, temperature is set to 0 for the consistent performance.

\end{document}